\documentclass[conference]{IEEEtran}

\usepackage{cite}
\usepackage{amsmath,amssymb,amsfonts}
\usepackage{graphicx}
\usepackage{textcomp}
\usepackage{xcolor}
\usepackage{hyperref}
\usepackage{booktabs}
\usepackage{tabularx}
\usepackage{float}
\usepackage{fancyhdr} 
\usepackage{placeins}

\def\BibTeX{{\rm B\kern-.05em{\sc i\kern-.025em b}\kern-.08em
    T\kern-.1667em\lower.7ex\hbox{E}\kern-.125emX}}

\fancypagestyle{firstpage}{%
  \fancyhf{}
}

\begin{document}

\title{A Lightweight CNN--Attention--BiLSTM Architecture for Multi-Class Arrhythmia Classification on Standard and Wearable ECGs}

\author{
\IEEEauthorblockN{Vamsikrishna Thota\IEEEauthorrefmark{1}, 
Hardik Prajapati\IEEEauthorrefmark{1}, 
Yuvraj Joshi\IEEEauthorrefmark{1}, 
Shubhangi Rathi\IEEEauthorrefmark{1}}
\IEEEauthorblockA{\IEEEauthorrefmark{1}Infocusp Innovations, India}
}

\maketitle

\begin{center}
    \vspace{-1em}
    {\small \textit{Accepted at CISP-BMEI 2025}}
\end{center}

\maketitle
\thispagestyle{firstpage}

\begin{abstract}
Early and accurate detection of cardiac arrhythmias is vital for timely diagnosis and intervention. We propose a lightweight deep learning model combining 1D Convolutional Neural Networks (CNN), attention mechanisms, and Bidirectional Long Short-Term Memory (BiLSTM) for classifying arrhythmias from both 12-lead and single-lead ECGs. Evaluated on the CPSC 2018 dataset, the model addresses class imbalance using a class-weighted loss and demonstrates superior accuracy and F1-scores over baseline models. With only 0.945 million parameters, our model is well-suited for real-time deployment in wearable health monitoring systems. The source code is available at \url{https://github.com/infocusp/tiny_arrhythmia_classification}.
\end{abstract}


\section{Introduction}
Cardiovascular diseases (CVDs) remain the leading cause of morbidity and mortality worldwide, accounting for approximately 17.9 million deaths each year according to the World Health Organization (WHO) \cite{who2021cvd}. Electrocardiography (ECG) is a simple yet effective technique for detecting arrhythmias. ECG is a widely adopted, non-invasive technique for monitoring the heart’s electrical activity, typically using a standard 12-lead configuration. This system involves the placement of 10 physical electrodes to produce 12 distinct leads, each offering a unique electrical perspective of the heart's function.
Specifically, the electrode setup includes four limb electrodes: right arm (RA), left arm (LA), right leg (RL), and left leg (LL) and six precordial chest electrodes, V1 through V6, strategically positioned on the body. The 12 leads are derived as follows:
Limb Leads: I (LA - RA), II (LL - RA), III (LL - LA)
Augmented Leads: aVR, aVL, aVF (each computed using one limb electrode against the average of the remaining two)
Chest Leads: V1 to V6, which record activity across the transverse plane.
This comprehensive spatial coverage enables precise identification of cardiac abnormalities, including arrhythmias\cite{Kligfield2007},\cite{goldberger2000physionet},\cite{sokolova2009systematic}.

Cardiac arrhythmias are disruptions in the normal rhythm of the heart and can cause serious cardiovascular complications if not detected early. ECGs are indispensable tools for identifying such conditions, making automated arrhythmia detection an important area of research. Traditional statistical or rule-based models often struggle with variability in signal morphology, noise, and class imbalance \cite{pan1985real}, \cite{sokolova2009systematic}.

In recent years, deep learning has revolutionized ECG classification. Convolutional Neural Networks (CNNs), in particular, have demonstrated strong performance by automatically extracting discriminative features from raw signals \cite{hannun2019cardiologist}, \cite{rajpurkar2017cardiologist}. However, ECG analysis with deep learning still presents challenges such as capturing long-term temporal dependencies, managing high intra-class variance, and addressing underrepresented classes in imbalanced datasets.

To mitigate these issues, hybrid architectures have been proposed, combining CNNs with recurrent layers like Long Short-Term Memory (LSTM) networks. These models excel at modeling temporal relationships in sequential data. For example,  \cite{sokolova2009systematic},\cite{yao2018time} showed significant improvements in classification by fusing CNNs with LSTM layers. Additionally, attention mechanisms have been incorporated to enhance focus on diagnostically relevant portions of the ECG waveform. Moreover, many of the state-of-the-art deep learning architectures \cite{yao2018time},\cite{Geng2023ECG}, \cite{ElGhaish2024ECGTransForm} are computationally intensive, making them less suitable for deployment on resource-constrained platforms such as wearable devices. Several lightweight architectures have been proposed in recent studies \cite{alamatsaz2022lightweight, jiang2024dceten}; however, these models were primarily evaluated on the MIT-BIH Arrhythmia Database, which contains limited variation in arrhythmia classes. To address these limitations, we introduce a lightweight architecture that balances accuracy and efficiency. Our model incorporates a streamlined attention mechanism to focus on clinically relevant patterns without introducing significant computational overhead.

For wearable devices, single-lead ECG is commonly used to detect arrhythmia-related problems \cite{niu2023diagnostic}, \cite{liu2024arrhythmia}. There is a study \cite{chen2020detection} related to single-lead experiments in which the authors also used the CPSC 2018 dataset, but with different leads. In our proposed work, we utilized the CPSC 2018 dataset for both standard 12-lead ECG and single-lead ECG, specifically using Lead I, which is well-suited for wearable devices.

The CPSC 2018 data set has become a standard benchmark for evaluating ECG classification models \cite{yildirim2018lstm}. It provides a diverse range of arrhythmia classes, yet remains affected by significant class imbalance. To counter this, techniques such as class-weighted loss functions and attention-based refinement have been utilized to improve generalization.
In this paper, we propose a lightweight yet powerful deep learning framework that integrates CNNs, attention mechanisms, and Bidirectional LSTM (BiLSTM) layers for multi-class arrhythmia classification. Our model captures both local features and long-term dependencies while maintaining computational efficiency. We address class imbalance through weighted loss functions and demonstrate superior performance compared to baseline models resent-18 \cite{ecg_resnet18} and TI-CNN \cite{yao2018time} on the CPSC 2018 dataset. The proposed architecture is optimized for practical deployment in both clinical and wearable health monitoring settings.

The rest of this paper is organized as follows: Section~\ref{sec:dataset} describes the dataset used in this study. Section~\ref{sec:preparation} outlines the dataset preparation and preprocessing steps. Section~\ref{sec:methodology} details the proposed methodology. Section~\ref{sec:experiments} presents the experimental setup, and Section~\ref{sec:results} discusses the results and performance analysis. Finally, Section~\ref{sec:conclusion} concludes the paper.

\section{Dataset}
\label{sec:dataset}

This study utilizes the CPSC 2018 ECG dataset, publicly available on PhysioNet. It contains 12-lead ECG recordings annotated with nine clinically relevant rhythm classes: Atrial Fibrillation (AF), 1st Degree AV Block (IAVB), Left and Right Bundle Branch Blocks (LBBB, RBBB), Sinus Rhythm (SNR), ST Depression (STD), ST Elevation (STE), Premature Ventricular Contraction (PVC), and Premature Atrial Contraction (PAC). Class annotations are provided using SNOMED CT codes in the corresponding header (\texttt{.hea}) files.

Each recording consists of a pair of files: a \texttt{.hea} file containing metadata and diagnostic labels, and a \texttt{.mat} file storing the raw ECG signal. All recordings were originally sampled at 500 Hz.

The number of recordings per class and their durations vary significantly. Table~\ref{tab:merged-class-stats} presents a consolidated summary of class-wise sample counts and signal duration statistics.

\begin{table}[h]
\centering
\caption{Combined Statistics: Class Distribution and Signal Duration}
\label{tab:merged-class-stats}
\begin{tabular}{lccccc}
\hline
\textbf{Class} & \textbf{Count} & \textbf{Mean (s)} & \textbf{Std Dev} & \textbf{Min (s)} & \textbf{Max (s)} \\
\hline
AF   & 1221 & 15.07 & 8.73  & 9.0    & 74.0 \\
IAVB & 722  & 14.42 & 7.08  & 9.998  & 54.31 \\
LBBB & 199  & 15.10 & 8.10  & 9.0    & 65.0 \\
PAC  & 544  & 19.30 & 12.39 & 9.0    & 74.0 \\
PVC  & 627  & 20.84 & 15.39 & 6.0    & 144.0 \\
RBBB & 1675 & 14.73 & 9.00  & 9.998  & 118.0 \\
SNR  & 918  & 15.43 & 7.64  & 10.0   & 64.0 \\
STD  & 786  & 15.65 & 9.79  & 8.0    & 138.0 \\
STE  & 185  & 17.31 & 10.74 & 10.0   & 60.0 \\
\hline
\end{tabular}
\end{table}

\section{Data Preparation}
\label{sec:preparation}

To ensure consistency and facilitate model training, several preprocessing steps were applied to the raw ECG signals:

\begin{itemize}
    \item \textbf{Downsampling:} All signals were resampled from 500 Hz to 250 Hz to reduce data dimensionality and computational overhead.
    
    \item \textbf{Length Normalization:} Each signal was standardized to a fixed duration of 60 seconds (15,000 samples at 250 Hz). Recordings longer than 60 seconds were truncated, while shorter ones were zero-padded.
    
    \item \textbf{Channel Structure:} In the multi-lead setup, each sample was represented as a 2D array of dimensions $12 \times 15000$. For single-lead analysis, only Lead I was retained, forming arrays of size $1 \times 15000$.
    
    \item \textbf{Filtering:} A high-pass filter was applied to remove baseline wander and low-frequency noise components.
\end{itemize}

The dataset was divided into training (80\%), validation (10\%), and testing (10\%) sets. Additionally, 10-fold cross-validation was performed to evaluate the robustness of the model. A custom data generator handled batch-wise loading with a batch size of 32, providing normalized ECG arrays and one-hot encoded labels for training.

\section{Methodology}
\label{sec:methodology}
\begin{figure}[H]
    \centering
    \includegraphics[width=0.4\textwidth]{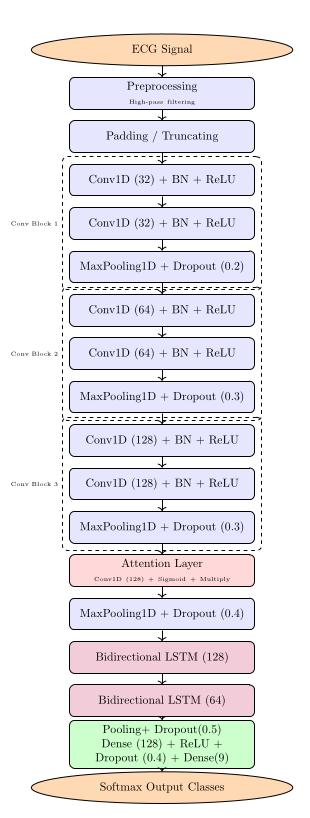}
    \caption{Proposed hybrid deep learning architecture for ECG-based arrhythmia classification.}
    \label{fig:model-architecture}
\end{figure}

\FloatBarrier
The proposed model architecture, shown in Fig.~\ref{fig:model-architecture}, is a deep hybrid network that combines enhanced convolutional processing, an attention mechanism, and recurrent sequence modeling to effectively classify arrhythmias from raw ECG signals.

The architecture begins with three sequential convolutional blocks, each consisting of two Conv1D layers with ReLU activation, batch normalization, max pooling, and dropout. These blocks progressively extract higher-level temporal features while reducing the dimensionality of the input signal.

Following the convolutional layers, a lightweight attention mechanism is applied via a Conv1D layer with a sigmoid activation, producing an attention map. This map is multiplied element-wise with the feature map to emphasize salient temporal patterns.

The output is then fed into two stacked Bidirectional LSTM layers. These layers are designed to capture long-range temporal dependencies in both forward and backward directions, which are essential for identifying rhythm-related abnormalities that span longer time windows.

A global average pooling layer reduces the temporal se-
quence to a feature vector, followed by dense layers for classi-
fication. The final softmax layer produces class probabilities.
Regularization is applied throughout using L2 penalties and
dropout to mitigate overfitting. The model is compiled with
the Adam optimizer and supports gradient clipping for stable
training.

A global average pooling layer reduces the temporal sequence to a feature vector, followed by dense layers for classification. The final softmax layer produces class probabilities. Regularization is applied throughout using L2 penalties and dropout to mitigate overfitting. The model is compiled with the Adam optimizer and supports gradient clipping for stable training.

Let the input ECG signal be denoted as:
\begin{equation}
X \in \mathbb{R}^{C \times T}
\label{eq:input}
\end{equation}

where:
\begin{itemize}
    \item $C$ is the number of channels (e.g., 12 for 12-lead or 1 for single-lead ECG),
    \item $T$ is the number of time steps (signal length).
\end{itemize}

\subsection{Convolutional Feature Extraction}

We employ a stack of 1D convolutional layers to extract temporal and morphological features from the ECG signals. The first convolution transforms the raw input $X$ into local feature maps $F^{(1)}$, which are refined into higher-level features $F^{(2)}$ by a second convolution. Each convolution is followed by Batch Normalization and ReLU activation to stabilize learning and introduce non-linearity. Max-pooling reduces the temporal dimension, and dropout is applied for regularization. The process is described in Equation~\ref{eq:conv}.

\begin{align}
F^{(1)} &= \text{ReLU}(\text{BN}(\text{Conv1D}_1(X))) \nonumber \\
F^{(2)} &= \text{ReLU}(\text{BN}(\text{Conv1D}_2(F^{(1)}))) \nonumber \\
F^{(2)} &= \text{Dropout}(\text{MaxPool}(F^{(2)}))
\label{eq:conv}
\end{align}

Where:
\begin{itemize}
    \item $\text{Conv1D}_i$: $i$-th 1D convolutional layer,
    \item $\text{BN}$: Batch Normalization,
    \item $\text{ReLU}$: Activation function,
    \item $\text{MaxPool}$: Temporal downsampling,
    \item $\text{Dropout}$: Regularization to prevent overfitting.
\end{itemize}

The output feature map:
\begin{equation}
F^{(2)} \in \mathbb{R}^{C' \times T'}
\label{eq:conv_out}
\end{equation}
has reduced temporal length $T'$ and increased channel depth $C'$.

\subsection{Attention Mechanism}
To enhance the discriminative power of the learned temporal features, an attention mechanism is employed between convolution and BiLSTM. The mechanism assigns weights to different time steps, enabling the model to focus on informative parts of the signal.

\begin{align}
A &= \sigma(\text{Conv1D}_{\text{att}}(F^{(2)})) \nonumber \\
F^{\text{att}} &= F^{(2)} \odot A
\label{eq:attention}
\end{align}

Where:
\begin{itemize}
    \item $\text{Conv1D}_{\text{att}}$ is a $1 \times 1$ convolution layer that reduces the feature dimension,
    \item $\sigma$ is the sigmoid activation function,
    \item $A \in \mathbb{R}^{C' \times T'}$ is the attention map that highlights salient temporal regions,
    \item $\odot$ denotes element-wise multiplication,
    \item $F^{\text{att}}$ is the resulting attention-weighted feature representation.
\end{itemize}

As shown in Equation~\ref{eq:attention}, the attention module adaptively scales each time step of the feature map, thereby improving the model’s sensitivity to arrhythmia-specific signal patterns.

\subsection{Bidirectional LSTM Layers}
\begin{equation}
H^{(1)} = \text{BiLSTM}_1(F^{\text{att}}), \quad H^{(2)} = \text{BiLSTM}_2(H^{(1)})
\label{eq:bilstm}
\end{equation}

Where:
\begin{itemize}
    \item $\text{BiLSTM}_i$ processes the sequence in both forward and backward directions,
    \item $H^{(2)} \in \mathbb{R}^{2d \times T''}$, with $d$ being the LSTM hidden size per direction.
\end{itemize}

\subsection{Global Average Pooling and Classification}
\begin{align}
z &= \text{GAP}(H^{(2)}) \nonumber \\
z &= \text{ReLU}(W_1 z + b_1), \quad z = \text{Dropout}(z) \nonumber \\
\hat{y} &= \text{softmax}(W_2 z + b_2)
\label{eq:classifier}
\end{align}

Where:
\begin{itemize}
    \item $W_1, W_2$ are learnable weight matrices,
    \item $b_1, b_2$ are the associated biases,
    \item $\hat{y} \in \mathbb{R}^K$ is the probability vector over $K$ arrhythmia classes.
\end{itemize}

\subsection{Loss Function}
We used a weighted cross-entropy loss to address class imbalance. This approach assigns higher weights to underrepresented classes and lower weights to overrepresented classes, ensuring a more balanced contribution to the overall loss.
\begin{equation}
\mathcal{L} = - \sum_{i=1}^{K} w_i \cdot y_i \cdot \log(\hat{y}_i)
\label{eq:loss}
\end{equation}

Where:
\begin{itemize}
    \item $K$ is the total number of classes,
    \item $y_i \in \{0, 1\}$ is the one-hot encoded true label for class $i$,
    \item $\hat{y}_i \in [0, 1]$ is the predicted probability for class $i$,
    \item $w_i$ is the weight assigned to class $i$.
\end{itemize}


\section{Experiments}
\label{sec:experiments}
We conducted experiments using both 12-lead standard ECG signals and single Lead-I ECG signals to evaluate the performance of our model. The results of these experiments are presented and discussed in the following section. For comparison, we considered two baseline models: ResNet-18~\cite{ecg_resnet18} and TI-CNN~\cite{yao2018time}. Additionally, we included another baseline—an ECG classification method based on multi-task learning and CoT attention mechanism~\cite{Geng2023ECG}—for 10-fold cross-validation comparison.

The proposed model was trained and tested on a server with a Intel(R) Xeon(R) E5-2623 v3 CPU @ 3.00GHz, 128 GB memory, and NVIDIA TITAN Xp 12GB GPU. We used Ubuntu 24.04.2 LTS OS and the model was implemented with tensorflow 2.18.0.

\textbf{Training Strategy:} The model is trained using the Adam optimizer with an initial learning rate of 0.001 and L2 regulization with mulication factor of 0.001. A learning rate scheduler reduces the rate by half every 20 epochs. To handle the class imbalance inherent in the CPSC 2018 dataset, class weights are computed and used during training. Sparse categorical weighted cross-entropy loss is employed to improve class imbalance and generalization.

The training pipeline is integrated with MLflow for full experiment tracking, including parameter logging, training metrics, artifact saving (e.g., confusion matrix, precision-recall curves), and model checkpoints. Additional callbacks such as early stopping, learning rate reduction on plateau, and best model checkpointing are used to ensure optimal performance and prevent overfitting.

\FloatBarrier
\section{Results and Discussion}
\label{sec:results}

In this work, we used standard classification metrics: Precision, Recall, F1-Score, and Area Under the ROC Curve (AUC) for each class. 
\begin{equation}
\text{Precision} = \frac{\text{TP}}{\text{TP} + \text{FP}}
\label{eq:precision}
\end{equation}

\begin{equation}
\text{Recall} = \frac{\text{TP}}{\text{TP} + \text{FN}}
\label{eq:recall}
\end{equation}

\begin{equation}
\text{F1-Score} = \frac{2 \times \text{Precision} \times \text{Recall}}{\text{Precision} + \text{Recall}}
\label{eq:f1}
\end{equation}

Where:
\begin{itemize}
    \item TP (True Positives): Correctly predicted positives,
    \item FP (False Positives): Incorrectly predicted positives,
    \item FN (False Negatives): Actual positives that the model missed.
\end{itemize}

\FloatBarrier

\subsection{12-Lead ECG Performance}
\begin{figure}[h]
    \centering
    \includegraphics[width=1\linewidth]{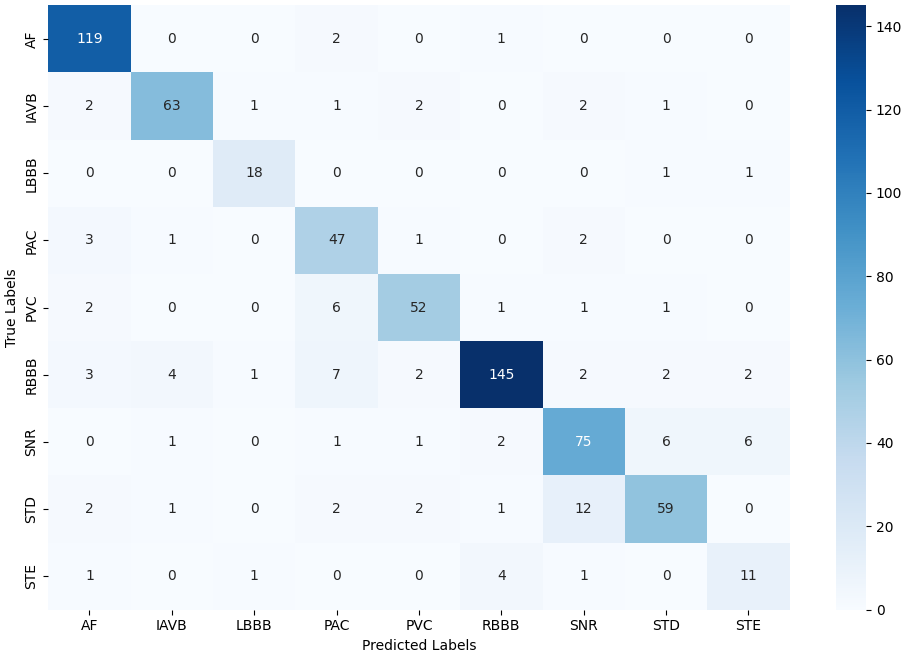}
    \caption{Confusion matrix of our proposed model on the 12-lead ECG classification task.}
    \label{fig:conf_matrix}
\end{figure}

In this section, we present the performance of proposed model on the 12-lead ECG data (see Table~\ref{tab:classification_performance}). The corresponding confusion matrix in Figure~\ref{fig:conf_matrix} further illustrates the model’s effectiveness in distinguishing between various arrhythmia classes.

\begin{table}[H]
\centering
\caption{Classification Performance for 12-lead ECG}
\label{tab:classification_performance}
\renewcommand{\arraystretch}{2} 
\footnotesize 
\resizebox{0.98\columnwidth}{!}{
\begin{tabular}{llllll}
\hline
\textbf{Class} & \textbf{TI-CNN (F1)} & \textbf{ResNet-18 (F1)} & \textbf{Ours (F1)} & \textbf{Support} & \textbf{AUC (Ours)} \\
\hline
SNR   & 0.759 & 0.77 & \textbf{0.80} & 92  & 0.9692 \\
AF    & 0.807 & 0.92 & \textbf{0.94} & 122 & 0.9974 \\
I-AVB & 0.855 & 0.87 & \textbf{0.89} & 72  & 0.9780 \\
LBBB  & 0.844 & 0.86 & \textbf{0.88} & 20  & 0.9982 \\
RBBB  & 0.837 & \textbf{0.92} & 0.90 & 168 & 0.9888 \\
PAC   & 0.606 & 0.76 & \textbf{0.78} & 54  & 0.9672 \\
PVC   & 0.712 & 0.81 & \textbf{0.85} & 63  & 0.9722 \\
STD   & 0.742 & 0.79 & \textbf{0.79} & 79  & 0.9635 \\
STE   & 0.556 & 0.55 & \textbf{0.58} & 18  & 0.8918 \\
\hline
\textbf{Average Precision} & 0.775 & 0.86 & \textbf{0.86} & 688 & \textbf{Avg AUC: 0.9692} \\
\textbf{Average Recall}    & 0.775 & 0.85 & \textbf{0.86} & 688 & \\
\textbf{Average F1-Score}  & 0.773 & 0.79 & \textbf{0.86} & 688 & \\
\hline
\end{tabular}%
}
\end{table}

\subsection{10-Fold Cross-Validation Comparison 12-Lead ECG}
To further evaluate the robustness of our proposed model, we conducted a 10-fold cross-validation experiment and compared its performance with a recent baseline: Geng et al. ~\cite{Geng2023ECG}. 

Table~\ref{tab:crossval_comparison} summarizes the average F1-Score across the 10 folds. Our model achieved a slightly better score, indicating improved generalization capability over the baseline.

\begin{table}[H]
\centering
\caption{10-Fold Cross-Validation F1-Score Comparison}
\label{tab:crossval_comparison}
\renewcommand{\arraystretch}{1.5}
\footnotesize
\begin{tabular}{lcc}
\hline
\textbf{Method} & \textbf{F1-Score} \\
\hline
Geng et al.~\cite{Geng2023ECG} & 0.827 \\
\textbf{Ours} & \textbf{0.839} \\
\hline
\end{tabular}
\end{table}

This result demonstrates that our model performs competitively and generalizes slightly better across different data splits.

\FloatBarrier

\subsection{Single-Lead (Lead-I) Performance}

\begin{figure}[h]
    \centering
    \includegraphics[width=1\linewidth]{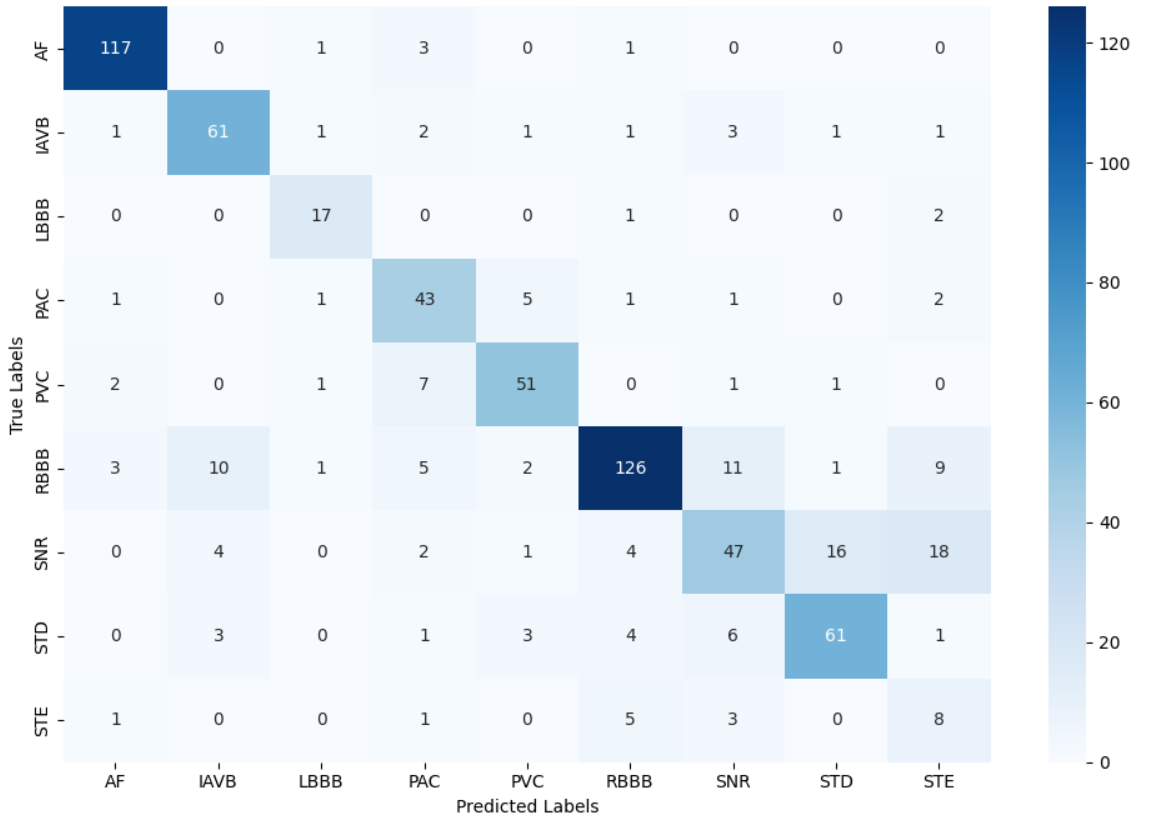}
    \caption{Confusion matrix of our proposed model on the single-lead (Lead-I) ECG classification task.}
    \label{fig:conf_matrix_singlelead}
\end{figure}

To evaluate the model's robustness in resource-constrained settings, we also trained and tested on single-lead (Lead-I) ECG signals. Table~\ref{tab:singlelead_performance} presents classification results using the same metrics. In addition, the confusion matrix in Figure~\ref{fig:conf_matrix_singlelead} illustrates the model's ability to differentiate between arrhythmia classes.

While the performance on single-lead (Lead-I) ECG was lower than that of the 12-lead configuration, this evaluation was conducted to assess the model’s robustness and generalizability in low-resource settings, such as wearable or portable ECG monitoring devices where only a single lead is available.

\begin{table}[h]
\centering
\caption{Classification Performance for Single-Lead (Lead-I) ECG}
\label{tab:singlelead_performance}
\renewcommand{\arraystretch}{1.6}
\resizebox{\columnwidth}{!}{%
\begin{tabular}{llllll}
\hline

\textbf{Class} & \textbf{ResNet-18 (F1)} & \textbf{Ours (F1)} & \textbf{Support} & \textbf{AUC (Ours)} \\
\hline
SNR   & \textbf{0.69}    & 0.57 & 92  & 0.9467 \\
AF    & 0.94  & \textbf{0.95} & 122 & 0.9967 \\
I-AVB & \textbf{0.90}   & 0.81 & 72  & 0.9809 \\
LBBB  & \textbf{0.82}   & 0.81 & 20  & 0.9923 \\
RBBB  & \textbf{0.84}  & 0.81 & 168 & 0.9668 \\
PAC   & 0.67   & \textbf{0.73} & 54  & 0.9689 \\
PVC   & 0.77    & \textbf{0.81} & 63  & 0.9783 \\
STD   & 0.71  &  \textbf{0.77} & 79  & 0.9568 \\
STE   & 0.23   & \textbf{0.27} & 28  & 0.8510 \\
\hline
\textbf{Average Precision} & \textbf{0.80}   & 0.79 & 688 & \textbf{Avg AUC: 0.9598}\\
\textbf{Average Recall}    & \textbf{0.78}    & 0.77 & 688 & \\
\textbf{Average F1-Score}  & \textbf{0.79}   & 0.78 & 688 &  \\
\hline
\end{tabular}%
}
\end{table}
\FloatBarrier

\subsection{Model Parameters Comparison}

Table~\ref{tab:model_comparison} provides a comparison of the number of parameters for different models used in this study. As shown, our model achieves a significantly lower parameter count compared to both TI-CNN and ResNet-18, making it highly efficient for real-time or edge deployment.

\begin{table}[t]
\centering
\caption{Comparison of Model Parameters}
\label{tab:model_comparison}
\renewcommand{\arraystretch}{1.3} 
\scalebox{1}{ 
\begin{tabular}{|l|l|}
\hline
\textbf{Model} & \textbf{Total Parameters} \\
\hline
TI-CNN~\cite{yao2018time} & 5.25M \\
ResNet-18 (1D)~\cite{ecg_resnet18} & 8.76M \\
Geng et al.~\cite{Geng2023ECG} & 1.10M \\
\textbf{Ours (Proposed)} & \textbf{0.945M} \\
\hline
\end{tabular}
}
\end{table}

Our model stands out with the lowest number of parameters (0.945M), making it lightweight and ready for deployment in resource-constrained environments. Despite its smaller size, it leverages BiLSTM and attention mechanisms, enabling it to capture temporal dependencies and extract the most relevant features from the ECG signals.

\begin{itemize}
    \item \textbf{TI-CNN}~\cite{yao2018time}: With 5.25M parameters, TI-CNN has more layers and complexity, which can lead to higher computational cost and longer inference times, particularly in real-time applications.
    
    \item \textbf{ResNet-18}~\cite{ecg_resnet18}: With 8.76M parameters, ResNet-18 is the most parameter-heavy among the compared models. While it performs well, its large size and computational demands make it less suitable for edge deployment or resource-constrained environments.
    
    \item \textbf{Geng et al. (CoT + MTL)}~\cite{Geng2023ECG}: This method leverages a multi-task learning framework and CoT attention mechanism, achieving strong performance (F1-score: 0.827) under 10-fold cross-validation. However, it may involve higher architectural complexity and training time due to the multi-branch design.
    
    \item \textbf{Our Model}: Our proposed model strikes an optimal balance between accuracy and efficiency, achieving the highest F1-score (0.839) in the 10-fold cross-validation setup. With fewer parameters and lower computational requirements, it is well-suited for real-time and edge device deployment.
\end{itemize}

\FloatBarrier
\subsection{Edge Deployment and Inference Efficiency}

To assess the deployment feasibility of our proposed model in low-power and real-time environments, we tested the  TensorFlow Lite (TFLite) version on a Raspberry Pi 4 Model B (4GB RAM). The evaluation confirms that our model is both lightweight and fast enough for edge applications. The inference performance metrics are summarized below:

\begin{itemize}
    \item \textbf{Model Format:} TensorFlow Lite
    \item \textbf{Model Size:} 3.66 MB
    \item \textbf{Platform:} Raspberry Pi 4 Model B (4GB RAM)
    \item \textbf{Inference Runs:} 100
    \item \textbf{Total Inference Time:} 17.41 seconds
    \item \textbf{Average Inference Time per Sample:} 0.174 seconds
    \item \textbf{RAM Before Inference:} 443.74 MB
    \item \textbf{RAM After Inference:} 453.84 MB
    \item \textbf{Estimated RAM Used During Inference:} ~10.1 MB
    \item \textbf{Peak Tracemalloc Memory Usage:} 0.00 MB
\end{itemize}

These results validate the suitability of our model for deployment on resource-constrained edge devices, such as wearable ECG monitors. The compact model size and sub-200 ms average inference time demonstrate its potential for real-time arrhythmia classification in portable healthcare solutions.

\FloatBarrier
\subsection{Ablation Study}

To evaluate the contribution of each architectural component in our proposed model, we conducted an ablation study by progressively integrating attention and BiLSTM layers. The results in Table~\ref{tab:ablation_study} demonstrate the performance gains (measured via average F1-score) and parameter costs for each configuration.

\begin{table}[h]
\centering
\caption{Ablation Study: Impact of Attention and BiLSTM on Model Performance}
\label{tab:ablation_study}
\renewcommand{\arraystretch}{1.5}
\resizebox{0.9\columnwidth}{!}{
\begin{tabular}{|l|c|c|}
\hline
\textbf{Model Configuration} & \textbf{Avg. F1-Score (12-lead)} & \textbf{Parameters} \\
\hline
CNN Only & 0.81 & 0.52M \\
CNN + Attention & 0.84 & 0.74M \\
CNN+ BiLSTM &0.85 & 0.929M \\
CNN + Attention + BiLSTM (Proposed) & \textbf{0.86} & \textbf{0.945M} \\
\hline
\end{tabular}
}
\end{table}

As shown, adding the attention mechanism improves the model's ability to focus on relevant signal regions, enhancing the F1-score by 3\%. Incorporating BiLSTM further boosts the performance by modeling temporal dependencies in the ECG signal, achieving the best F1-score of 0.86 with a modest increase in parameter count.

Thus, while TI-CNN and ResNet-18 provide competitive performance, our proposed model is significantly more lightweight and efficient, making it the best choice for real-time or edge device deployment.


\section{Conclusion}
\label{sec:conclusion}
In this paper, we proposed a lightweight deep learning architecture combining CNN, attention mechanisms, and BiLSTM  for multi-class arrhythmia classification using both 12-lead and single-lead ECG recordings. Our approach addresses key challenges such as temporal dependency modeling, class imbalance, and overfitting while maintaining a low parameter count suitable for deployment on resource-constrained devices. Experimental results on the CPSC 2018 dataset demonstrate that our model achieves competitive performance compared to state-of-the-art baselines. In future work, we aim to explore generalization across other ECG datasets, extend support to real-time streaming inputs, and integrate model optimization techniques for embedded platforms.

\bibliographystyle{IEEEtran}
\bibliography{refs}

\end{document}